# Calculating Uncertainty Intervals From Conditional Convex Sets of Probabilities

Serafín Moral
Departamento de Ciencias de la Computación e I.A.
Universidad de Granada
18071 - Granada - Spain
e-mail: smoral@ugr.es

## Abstract

In Moral, Campos (1991) and Cano, Moral, Verdegay-López (1991) a new method of conditioning convex sets of probabilities has been proposed. The result of it is a convex set of non-necessarily normalized probability distributions. The normalizing factor of each probability distribution is interpreted as the possibility assigned to it by the conditioning information. From this, it is deduced that the natural value for the conditional probability of an event is a possibility distribution. The aim of this paper is to study methods of transforming this possibility distribution into a probability (or uncertainty) interval. These methods will be based on the use of Sugeno and Choquet integrals. Their behaviour will be compared in basis to some selected examples.

## 1 INTRODUCTION

Dempster (1967) introduced two different ways of conditioning that can be applied in the Theory of Evidence (see also Shafer, 1976). If $U$ is a finite set, $m$ is a basic probability assignment defined on it, and $(Bel, Pl)$ is the corresponding pair of belief-plausibility measures, then the first definition is based on a direct generalization of probability conditioning formula for the plausibility measure:

$$Pl_1(A|B) = \frac{Pl(A \cap B)}{Pl(B)}, \text{ if } Pl(B) \neq 0 \quad (1)$$

The corresponding expression for $Bel$ is not so simple:

$$Bel_1(A|B) = \frac{Pl(B) - Pl(\overline{A} \cap B)}{Pl(B)}, \text{ if } Pl(B) \neq 0 \quad (2)$$

The other definition was given in terms of the underlying probability distributions associated with a pair $(Bel, Pl)$:

$$\mathcal{P} = \{p \mid Bel(A) \leq P(A) \leq Pl(A), \forall A \subseteq U\} \quad (3)$$

**Remark.** If $p$ is a probability distribution, its associate probability measure will be denoted using the capital letter: $P$.

To calculate the conditional information with respect to a set $B$, all the probabilities are conditioned and the set $\mathcal{P}_B$ is calculated, where

$$\mathcal{P}_B = \{p(.|B) \mid p \in \mathcal{P}\} \quad (4)$$

The values of the corresponding plausibilities and beliefs are calculated as

$$Pl_2(A|B) = Sup\{P(A)|p \in \mathcal{P}_B\} = $$
$$= Sup\{P(A|B) \mid p \in \mathcal{P}\} \quad (5)$$

$$Bel_2(A|B) = Inf\{P(A)|p \in \mathcal{P}_B\} = $$
$$= Inf\{P(A|B) \mid p \in \mathcal{P}\} \quad (6)$$

In Demspter (1967) it is not said whether $(Bel_2(.|B), Pl_2(.|B))$ is a pair of belief-plausibility functions, that is, it has an associated basic probability assignment. In Campos, Lamata, Moral (1990), it was shown that $(Bel_2(.|B), Pl_2(.|B))$ may be expressed in the following way,

$$Pl_2(A|B) = \frac{Pl(A \cap B)}{Pl(A \cap B) + Bel(B - A)}$$
$$Bel_2(A|B) = \frac{Bel(A \cap B)}{Bel(A \cap B) + Pl(B - A)} \quad (7)$$

In Jaffray (1990) and Fagin, Halpern (1990), it is shown that $(Bel_2(.|B), Pl_2(.|B))$ is always a pair of belief-plausibility functions.

These two conditional beliefs are related. We have that for every $A, B \subseteq U$, the following relation is verified,

$$Bel_2(A|B) \leq Bel_1(A|B) \leq Pl_1(A|B) \leq Pl_2(A|B) \quad (8)$$

that is,

$$[Bel_1(A|B), Pl_1(A|B)] \subseteq [Bel_2(A|B), Pl_2(A|B)] \quad (9)$$

If we consider the upper and lower probabilities interpretation of belief functions, above intervals are interpreted as the set of possible values for an unknown probability value. From, this point of view, we can say that the first conditioning is more informative than the second: it produces smaller intervals. Now, the question is: Which conditioning is more appropriate for upper and lower probabilities?. In this paper we shall show that none of the two is (see also Moral, Campos (1991)). $(Bel_1(.|B), Pl_1(.|B))$ is too informative and $(Bel_2(.|B), Pl_2(.|B))$ too uninformative. Then we shall give a new method of conditioning proposed in Moral, Campos (1991) and Cano, Moral, Verdegay-López (1991) intermediate between this two and that, from our point of view, gives the right answer to the problem. This conditioning assigns to each event a possibility distribution (see Zadeh, 1978; Dubois, Prade, 1988) instead of an interval. In Moral, Campos (1991) and Cano, Moral, Verdegay-López (1991) was given a method of transforming this possibility distribution into an interval. The reason of this transformation is that for the final user of a system handling upper and lower probabilities it may be difficult to understand the result in terms of a Possibility Distribution or, at least, it may help to have an interval at the same time that the Possibility Distribution. In this paper we are going to criticize this method showing that it produces too short intervals. Then we shall propose new methods and compare them.

The organization of the paper is as follows: In the second section we introduce the mathematical elements used to represent upper and lower probabilities and then we give some prototipical examples, in order to compare the different conditioning methods. In the third section we study the new conditioning method assigning a possibility distribution to each event. In the fourth section we study the different methods of transforming this possibility distribution in probability intervals. Finally in the last section we discuss the content of the paper and give the conclusions.

## 2 PREVIOUS CONCEPTS AND PROTOTYPICAL EXAMPLES

Let $U$ be a finite set and $X$ a variable taking values on it. As representation of knowledge about how $X$ takes values on $U$, we shall consider convex sets of probability distributions. This representation is more general than those associated with belief functions, by means of expression (3). However if we take supremum and infimum of a convex set of probabilities we do not always obtain a pair of plausibility-belief functions. Furthermore, different convex sets may produce the same system of probability intervals. The use of convex sets of probabilities is also very well justified from a betting behaviour point of view, similar to the one used to justify bayesian probabilities, but less restrictive (see Walley, 1991).

With respect to the conditioning problem, the second definition is directly appliable to convex sets of probabilities: If we have a convex set of probability distributions on $U$, $\mathcal{P}$, then to condition to $B$ is equivalent to calculate

$$\mathcal{P}_B = \{p(.|B) \mid p \in \mathcal{P}\} \quad (10)$$

However the first definition of conditioning (see expressions (1,2)) is not directly appliable. To generalize it we can take as basis the fact that the convex set associated with $(Bel_1(.|B), Pl_1(.|B))$ is equal to (see Moral, Campos, 1991):

$$\mathcal{P}'_B = \{p(.|B) \mid p \in \mathcal{P} \text{ and } P(B) = Sup_{p_i \in \mathcal{P}} P_i(B)\} \quad (11)$$

That is we only consider the probabilities with a maximum value for the probability of the conditioning set $B$. Having expressed this conditioning in terms of convex sets, we shall consider that the first conditioning is equivalent to transform a convex set $\mathcal{P}$ into $\mathcal{P}'_B$, as in expression (11).

In the following we introduce two examples to show that none of this two conditioning formulas is very appropriate for upper and lower probabilities.

**Example 1** *Let us consider two urns, $u_1$ and $u_2$. $u_1$ has 999 red balls and 1 white. $u_2$ has 1 red ball, and 999 white. Assume an experiment consisting in selecting one of the urns by means of an unknown procedure and then picking up randomly a ball from that urn. What we want to test with this example is the abduction capability of a given conditioning procedure, more concretely, to determine what says the colour of the ball about the selected urn. It is clear that is the colour of the ball is red (r) the urn should be $u_1$ and if the color is white (w), $u_2$. In each case is the urn which better explains the observation. Let us see how the conditioning formulas work.*

*The set of possible probabilities associated with this example can be represented by the convex set generated by the two extreme probabilities:*

$$\begin{array}{ll} p_1(u_1, r) = 0.999, & p_1(u_1, w) = 0.001, \\ p_1(u_2, r) = 0.0, & p_1(u_2, w) = 0.0 \end{array} \quad (12)$$

$$p_2(u_1,r) = 0.0, \quad p_2(u_1,w) = 0.0,$$
$$p_2(u_2,r) = 0.001, \quad p_2(u_2,w) = 0.999 \quad (13)$$

*The first probability corresponds to the selection of the first urn. The second probability corresponds to the second urn.*

*Assume that we observe that the colour of the ball is red. The conditional probabilities are*

$$p_1(u_1,r|r) = 1.0, \quad p_1(u_1,w|r) = 0.0,$$
$$p_1(u_2,r|r) = 0.0, \quad p_1(u_2,w|r) = 0.0 \quad (14)$$

$$p_2(u_1,r|r) = 0.0, \quad p_2(u_1,w|r) = 0.0,$$
$$p_2(u_2,r|r) = 1.0, \quad p_2(u_2,w|r) = 0.0 \quad (15)$$

*The first conditioning is equivalent to consider the convex set generated by the conditional distributions $p_i(.|r)$, for which $P_i(r)$ is maximum: In this case, $p_1(.|r)$. The second conditioning is carried out by considering all the conditional probabilities, $p_1(.|r)$ and $p_2(.|r)$.*

*If we calculate probability intervals for $u_1$ and $u_2$, by taking supremum and infimum, we get*

- First Conditioning
$$u_1 \longrightarrow [1,1], \quad u_2 \longrightarrow [0,0] \quad (16)$$

- Second Conditioning
$$u_1 \longrightarrow [0,1], \quad u_2 \longrightarrow [0,1] \quad (17)$$

*We observe that with the first conditioning we are sure that the urn is $u_1$. That is not true. $u_1$ is the urn that best explains the result, but the red ball could also come from $u_2$. On the contrary, with the second conditioning, nothing is deduced about which is the selected urn. The ignorance interval, $[0,1]$, is assigned to both urns. In this example we observe that the first conditioning is too strong and the second too weak.*

**Example 2** *It could be argued that the strange behaviour of the first conditioning on former example is due to the fact that the initial convex set does not correspond to a belief function. Here we shall start with a very simple belief function and show that the conditioning ia also too strong.*

*Assume that we have an urn with red (r), black (b) and white (w) balls and that we know that there are: 10 red, 10 white and 20 that may be red or blacks. A ball is selected randomly from this urn.*

*This information may be represented by a mass assignment with focal elements:*

$$m(\{r\}) = 0.25, \; m(\{w\}) = 0.25,$$
$$m(\{r,b\}) = 0.5 \quad (18)$$

*The associated convex set is determined by the following extreme probabilities,*

$$p_1(r) = 0.75, \quad p_1(w) = 0.25, \quad p_1(b) = 0.0 \quad (19)$$

$$p_2(r) = 0.25, \quad p_2(w) = 0.25, \quad p_1(b) = 0.5 \quad (20)$$

*If we observe that the colour is red or white, then the first conditioning provides the convex set generated by $p_2(.|\{r,w\})$ and the second conditioning the convex set generated by $p_1(.|\{r,w\})$ and $p_2(.|\{r,w\})$.*

*The first conditioning assigns exact values of probability for the colour of the ball (there is only one possible probability). These values are,*

$$p_1(r|\{r,w\}) = 0.75, \; p_1(w|\{r,w\}) = 0.25,$$
$$p_1(b|\{r,w\}) = 0.00 \quad (21)$$

*Observe that the assigned probabilities are the same that the ones that would be considered if somebody tell us that all the balls are red or white, that is, if we learn that the colour of the 20 balls that were red or black, is red. However, these two situations are different. In the initial one we know that a particular ball randomly selected from the urn is red or white. In the second case we learn that all the balls of the urn are red or white. The second piece of information is stronger than the first. However, they two produce the same results. This will not be considered a very desirable property and we shall ask to a conditioning procedure to discriminate between these two situations. Second conditioning do this distinction.*

## 3  CONDITIONAL PROBABILITIES WITH POSSIBILITY VALUES

In Campos, Moral (1991) and Cano, Delgado, Moral (1991) we have introduced a new method of calculating conditional information. The idea is very simple. If we have a convex set of probabilities, $\mathcal{P}$, defined on a finite set $U$, and $B$ is a subset from $U$, the conditional information is given by the set,

$$\mathcal{P}''_B = \{p.I_B \mid p \in \mathcal{P}\} \quad (22)$$

where $I_B$ is the characteristic function of set $B$ and $p.I_B$ is pointwise multiplication.

**Example 3** *In the same conditions of Example 1, the conditional information is the convex set generated by the points,*

$$p_1.I_r(u_1,r) = 0.999, \quad p_1.I_r(u_1,w) = 0.0,$$
$$p_1.I_r(u_2,r) = 0.0, \quad p_1.I_r(u_2,w) = 0.0 \quad (23)$$

$$p_2.I_r(u_1,r) = 0.0, \quad p_2.I_r(u_1,w) = 0.0,$$
$$p_2.I_r(u_2,r) = 0.001, \quad p_2.I_r(u_2,w) = 0.0 \quad (24)$$

**Example 4** *In the conditions of Example 2, the conditional information is the convex set generated by the mappings,*

$$p_1.I_{\{r,w\}}(r) = 0.75, \quad p_1.I_{\{r,w\}}(w) = 0.25,$$
$$p_1.I_{\{r,w\}}(b) = 0.0 \tag{25}$$

$$p_2.I_{\{r,w\}}(r) = 0.25, \quad p_2.I_{\{r,w\}}(w) = 0.25,$$
$$p_2.I_{\{r,w\}}(b) = 0.0 \tag{26}$$

The main problem of this definition is to know what is the meaning of the result of conditioning. The result is a set which elements are mappings defined on $U$ and taking non-negative real numbers. These mappings are not probability distributions: in general, they do not add 1. We could think of normalizing all of them, but then we would obtain the second conditioning. In this point, it is important to remark that the normalizing factor contains valuable information. In Example 3, the normalizing factor of $p_1.I_r$ is 0.999 and the normalizing factor of $p_2.I_r$ is 0.001. This tell us that the first probability (and therefore the first urn) is *more possible* than the second.

Following above interpretation, let $n(p_i.I_B)$ be the probability obtained from $p_i.I_B$ by normalizing, that is, $n(p_i.I_B) = p_i(.|B)$ and $f(p_i.I_B)$ is the normalizing factor. We interpret $\mathcal{P}''_B$ as a possibility distribution (Zadeh, 1978; Dubois, Prade, 1988) defined on the set of conditional probabilities,

$$\pi : \mathcal{P}_B \longrightarrow [0,1] \tag{27}$$

with the following values,

$$\pi(p(.|B)) = \frac{Sup\{f(p_i.I_B) \mid n(p_i.I_B) = p(.|B)\}}{Sup\{f(p_i.I_B) \mid p_i \in \mathcal{P}\}} \tag{28}$$

Now, the value of the conditional probability of an event is not a real value, neither a probability interval, but a possibility distribution defined on the interval $[0,1]$. According to possibility calculus, we define the probability of an event, $A$, as a possibility $\pi_{A|B}$ on $[0,1]$, with the values,

$$\pi_{A|B}(x) = Sup\{\pi(p(.|B)) \mid P(A|B) = x, p(.|B) \in \mathcal{P}_B\} \tag{29}$$

In Example 3, we have that the probability of $u_1$ may be 1 with possibility 1. It can be also 0, but with possibility 1/999. The possibility $\pi_{u_1|r}$ is given in Figure 1. $\pi_{u_2|r}$ is given in Figure 2.

It is clear that for $u_1$ and $u_2$ the same interval of probabilities is possible: $[0,1]$. In this aspect it is similar to the second conditioning. However, for $u_1$ the probabilities near to 1 are more possible, and for $u_2$ the most posible probabilities are those near to 0. In this sense, the observation of the colour of the ball says something about the selected urn: It is more possible $u_1$. Also, it is important to remark that the conditional information is not as strong as in the first conditioning.

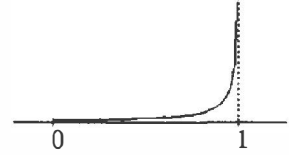

Figure 1: $\pi_{u_1|r}$ in Example 3

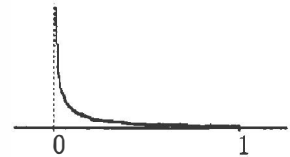

Figure 2: $\pi_{u_2|r}$ in Example 3

In Example 4, we observe that the set $\mathcal{P}_B$ has two extreme points: $p_1(.|\{r,w\})$ and $p_2(.|\{r,w\})$. But $\mathcal{P}''_B$ assigns the following possiblities to them:

$$\pi(p_1(.|\{r,w\})) = 1, \quad \pi(p_2(.|\{r,w\})) = 0.75 \tag{30}$$

This situation is different of the first conditioning: Now $p_1(.|\{r,w\})$ and $p_2(.|\{r,w\})$ are possible, and the situation is not the same that when we learn that all the balls are red or white. Also the result is something different of the second conditioning, $p_1(.|\{r,w\})$ is now more possible. The difference is not as great as in Example 3.

From our point of view, these conditional probabilities with associated possibilities contain all the relevant information. However, we think that its main problem is that sometimes they may be a bit complicated to be communicated to an ordinay user of a system working with upper and lower probabilities. Due to this, we shall consider methods of calculating an uncertainty interval for each event.

## 4 TRANSFORMATION OF A CONDITIONAL PROBABILITY ON AN INTERVAL

The basic elements for this transformation are Choquet's integral (Choquet, 1953/54) and Sugeno's integral (Sugeno, 1974). In the following we are going to give its definition for possibility and necessity measures.

If we have a possibility distribution $\pi$ defined on a set, $R$ (finite or infinite). Its associated possibility measure is a mapping,

$$\Pi : \wp(R) \longrightarrow [0,1] \quad (31)$$

given by,

$$\forall T \subseteq R, \Pi(T) = Sup\{\pi(t) \mid t \in T\} \quad (32)$$

The dual measure is called a necessity measure and is defined as

$$\forall T \subseteq R, N(T) = 1 - \Pi(\overline{T}) \quad (33)$$

If $g$ is a positive function defined on $R$, the Choquet integral of $g$ with respect to $\Pi$ is the value

$$I_C(g|\Pi) = \int_0^{+\infty} \Pi(\{x|g(x) \geq \alpha\})d\alpha \quad (34)$$

With respect to $N$, we have a similar expression,

$$I_C(g|N) = \int_0^{+\infty} N(\{x|g(x) \geq \alpha\})d\alpha \quad (35)$$

The Sugeno integral is defined for functions taking values on $[0,1]$. For possibility and necessity measures it has the following expressions,

$$\begin{aligned}I_S(g|\Pi) &= Sup_{\alpha \in [0,1]}\{\Pi(\{x|g(x) \geq \alpha\}) \wedge \alpha\} \\ &= Sup_x\{g(x) \wedge \pi(x)\}\end{aligned} \quad (36)$$

$$I_S(g|N) = Inf_{\alpha \in [0,1]}\{N(\{x|g(x) \geq \alpha\}) \vee \alpha\} \quad (37)$$

where $\wedge$ denotes the minimum and $\vee$ the maximum.

For functions $g$, taking values on $[0,1]$, it is inmediate to show that for every pair of measures, $\Pi$ and $N$, it is verified that,

$$I_C(g|\Pi) + I_C(1-g|N) = 1 \quad (38)$$

$$I_S(g|\Pi) + I_S(1-g|N) = 1 \quad (39)$$

$I_C(g|\Pi)$ is also called upper expectation of $g$ and $I_C(g|N)$ lower expectation (Dempster, 1967). These are the two integrals defined for non-additive measures. In this point we do not have any reason to choose one of them. For an axiomatic characterization of the properties underlying these two integrals see Campos, Lamata Moral (1991).

1. Rest_Possibility = 1
2. Possibility = 0
3. Upper_Choquet = 0
4. While Rest_Possibility > 0 do
   5. Select $p(.|B)$ from $Nextr(\mathcal{P}_B'')$ with maximum value of $P(A|B)$ among those verifying $\pi(p(.|B)) \geq$ Possibility
   6. Upper_Choquet = Upper_Choquet + $+(\pi(p(.|B)) -$ Possibility $) * P(A|B)$
   7. Rest_Possibility = Rest_Possibility $-(\pi(p(.|B)) -$ Possibility $)$
   8. Possibility $= \pi(p(.|B))$

Figure 3: Algorithm to calculate $P^*_C(A|B)$ (The value is Upper_Choquet).

In Moral, Campos (1991) we proposed the assignation of a conditional uncertainty interval to an event, $A$, by using Choquet's integral. The set $R$ was the set of normalized extreme points of the set $\mathcal{P}_B''$, that is, the conditional probabilities obtained from the extreme points of $\mathcal{P}_B''$. This set of extreme points will be called $Nextr(\mathcal{P}_B'')$. The possibility distribution is the one induced by equation (28). Function $g$ is defined by,

$$g(p(.|B)) = P(A|B) \quad (40)$$

In this way the upper uncertainty value is,

$$P^*_C(A|B) = I_C(g|\Pi) = I_C(P(A|B) \mid \Pi) \quad (41)$$

and the lower value,

$$P_{*C}(A|B) = I_C(g|N) = I_C(P(A|B) \mid N) \quad (42)$$

It is inmediate to show that $P_{*C}(A|B) \leq P^*_C(A|B)$ and by property (39), we have that

$$P_{*C}(\overline{A}|B) + P^*_C(A|B) = 1 \quad (43)$$

An algorithm to calculate $P^*_C(A|B)$ is given in Figure 3. This algorithm has a complexity of order $O(n^2)$, where $n$ is the number of points on $Nextr(\mathcal{P}_B'')$. If the probabilities in this set are ordered by decreassing value of $P(A|B)$ then the complexity is $O(n)$, but we have to take into account that the ordering algorithm takes a time of order $O(n.\log(n))$. A similar algorithm may be devised for $P_{*C}(A|B)$, but we may also calculate this value by means of equation (43).

**Example 5** *Following the case of Example 3, we have*

$$p^*_C(u_1|r) = 1 * 1 = 1$$

$$p^*_C(u_2|r) = 1 * (1/999) + 0 * (1 - 1/999) = 1/999$$

*In this way,*

$$[p_{*C}(u_1|r), p^*{}_C(u_1|r)] = [1 - 1/999, 1] \approx [0.999, 1]$$

*and*

$$[p_{*C}(u_2|r), p^*{}_C(u_2|r)] = [0, 1/999] \approx [0, 0.001]$$

*As we can see with these intervals, the colour of the ball gives relevant information about the selected urn, but not as much as the first conditioning.*

**Example 6** *Now, in the conditions of Example 4,*

$$[p_{*C}(r|\{r,w\}), p^*{}_C(r|\{r,w\})] =$$
$$= [0.5 * 0.5 + 0.5 * 0.75, 1 * 0.75] = [0.625, 0.75]$$

$$[p_{*C}(w|\{r,w\}), p^*{}_C(w|\{r,w\})] = [0.25, 0.375]$$

*In this example, we get also an intermediate interval between the intervals of the first and the second conditioning.*

In this point, it is important to remark that the resulting interval would have been different if we had considered as set $R$ the whole set $\mathcal{P}_B$, instead of $Nextr(\mathcal{P}''_B)$. To see that adding more probabilities from $\mathcal{P}_B$ the result chages, we shall consider, in the conditions of Example 3 and 5, the additional probability

$$p(u_1|r) = 0.5, \quad p(u_2|r) = 0.5 \quad (44)$$

This probability belongs to $\mathcal{P}_B$: it is a convex combination of $p_1(.|r)$ and $p_2(.|r)$. Its possibility may be calculated by means of expression (28), being equal to 0.002. Then aplying the algorithm of Figure 3, with this probability added to the extreme probabilities, we get the following intervals:

- For $u_1$ given $r$:
  $[1 - (1/999) - (0.002 - (1/999)) * 0.5, 1] \approx [0.9985, 1]$
- For $u_2$ given $r$:
  $[0, (1/999) + (0.002 - (1/999)) * 0.5] \approx [0, 0.0015]$

The interval has increased. We can verify that, in general, the consideration of the extreme points of $\mathcal{P}_B$ instead of the complete set reduces the interval. As the conditional set is $\mathcal{P}_B$ (with the associated possibilities), we can concluded that the intervals proposed in Moral, Campos (1991) and Cano, Delgado, Moral (1991) were too short. Furthermore, there is an important property that is not verified: continuity. Infinitesimal changes in the set $\mathcal{P}''_B$ can produce non infinitesimal changes on the intervals. If in example 3, we have the additional point,

$$\begin{array}{ll} p_3.I_r(u_1, r) = 0.00999 + \epsilon, & p_3.I_r(u_1, w) = 0.0, \\ p_3.I_r(u_2, r) = 0.00999 - \epsilon, & p_3.I_r(u_2, w) = 0.0 \end{array} \quad (45)$$

this point is also extreme. The associated normalized probability is,

$$p_3(u_1|r) = 0.5 + O(\epsilon), \quad p_3(u_2|r) = 0.5 + O(\epsilon) \quad (46)$$

and its possibility $0.002 + O(\epsilon)$. This probability distribution has to be considered and the intervals are the ones obtained by introducing probability (44) with a difference on the extremes of order $O(\epsilon)$. In short: very small changes in $\mathcal{P}''_B$ can produce new extreme points which produce a big change on the intervals.

To calculate the conditional intervals taking into account all the probabilities we need to do only some small changes in the algorithm of Figure 3. The new version includes a new variable, *Former_Probability*, initially set to 1, and that in each iteration contains the value of $P(A|B)$ in the previous iteration. Then, only sentence 6 is changed to:

6'. Upper_Choquet = Upper_Choquet +
+ (Possibility*Former_Probability $-\pi(p(.|B)) * P(A|B))$ +
+ Possibility*$\pi(p(.|B))$*(Former_Probability-$P(A|B))*$
$* \ln(\pi(p(.|B))/\text{Possibility})/(\text{Possibility} -\pi(p(.|B)))$

where ln denotes neperian logarithm. We shall not prove that this algorithms calculates the value of the integral, because the proof is only technical.

The complexity is the same than before ($O(n. \log(n))$) and now we do not have former continuity problems.

The values provided by the modified algorithm are always greater that the ones obtained with the previous one. So the intervals will be wider. In the case of Examples 3 and 5, the new intervals are,

$$[p'_{*C}(u_1|r), p^{*'}{}_C(u_1|r)] =$$
$$= [1 - \ln(999)/998, 1] \approx [0.9931, 1] \quad (47)$$

$$[p'_{*C}(u_2|r), p^{*'}{}_C(u_2|r)] =$$
$$= [0, \ln(999)/998] \approx [0, 0.0069] \quad (48)$$

The intervals have increased, but according to them we should follow thinking that the first urn was selected if the colour of the ball is red.

In the case of Examples 4 and 6, the intervals are

$$[p'_{*C}(r|\{r,w\}), p^{*'}{}_C(r|\{r,w\})] =$$
$$= [1 - 0.25(1 + \ln(2)), 0.75] \approx [0.5767, 0.75] \quad (49)$$

$$[p'_{*C}(w|\{r,w\}), p^{*'}{}_C(w|\{r,w\})] =$$
$$= [0.25, 0.25(1 + \ln(2))] \approx [0.25, 0.4233] \quad (50)$$

The result is also similar to the previously obtained but with a little wider intervals.

In the following we shall consider the results obtained by using Sugeno's integral. In this case,

$$P^*{}_S(A|B) = I_S(g|\Pi) = I_S(P(A|B) \mid \Pi) \qquad (51)$$

$$P_{*S}(A|B) = I_S(g|N) = I_S(P(A|B) \mid N) \qquad (52)$$

First, we shall consider as set $R$, the normalized extreme points of $\mathcal{P}_B''$: $Nextr(\mathcal{P}_B'')$. $P^*{}_S(A|B)$ can be easily calculated from equation (36) in linear time.

In Example 3 we get the following intervals,

- For $u_1$ given $r$: $[1 - (1/999), 1] \approx [0.999, 1]$
- For $u_2$ given $r$: $[0, (1/999)] \approx [0, 0.001]$

that is, the same results that using Choquet's integral.

In the case of Example 4, the intervals are:

- For $r$ given $\{r, w\}$: $[0.5, 0.75]$
- For $w$ given $\{r, w\}$: $[0.25, 0.5]$

In this case we obtain wider intervals that using Choquet's integral with all the points. However, this is not a fix rule. We do not always obtain wider or equal intervals than using Choquet's integral. If the initial probabilities of Example 2 had been,

$$p_1(r) = 1/3, \quad p_1(w) = 1/3, \quad p_1(b) = 1/3 \qquad (53)$$

$$p_2(r) = 0, \quad p_2(w) = 1/3, \quad p_1(b) = 2/3 \qquad (54)$$

and with the same conditioning set, the intervals calculated using Sugeno's integral would be:

- For $r$ given $\{r, w\}$: $[1/2, 1/2] = \{1/2\}$
- For $w$ given $\{r, w\}$: $[1/2, 1/2] = \{1/2\}$

The intervals calculated using the first conditioning are exactly the same, however the intervals for the second conditioning are:

- For $r$ given $\{r, w\}$: $[0, 1/2]$
- For $w$ given $\{r, w\}$: $[1/2, 1]$

In this case, the intervals are equal to the ones obtained with the first conditioning, too restrictive in comparison with the results of the second conditioning.

This irregular behaviour: sometimes less informative that the second conditioning and other times as informative as first conditioning is due to the consideration of $Nextr(\mathcal{P}_B'')$, instead of $\mathcal{P}_B$. If we consider the complete set we may dessign a procedure to calculate the upper value, $P^{*'}{}_S(A|B)$. Consider the set,

$$H = \{(t, r) \mid t = \pi(p(.|B)).P(A|B), r = \pi(p(.|B))\} \qquad (55)$$

and $Ext(H)$ the set of its extreme points. Let $S$ be the set

$$S = \{(u, v) \mid u = t/r, v = r, r \neq 0\} \qquad (56)$$

The algorithm to claculate $P^{*'}{}_S(A|B)$ follows the following steps:

1. Calculate the set $T$ of points $(u, v) \in S$ such that there is not point $(r, s) \in S$ verifying
$$r \geq u, s \geq v, (r, s) \neq (u, v)$$

2. If for every $(u, v) \in T$, is $u \geq v$, $P^{*'}{}_S(A|B)$ is the value
$$Sup\{v \mid \exists u, (u, v) \in T\}$$

3. If for every $(u, v) \in T$, is $u \leq v$, $P^{*'}{}_S(A|B)$ is the value
$$Sup\{u \mid \exists v, (u, v) \in T\}$$

4. If the conditions of steps 2 and 3 are not verified, let
$$u_0 = Sup\{u \mid (u, v) \in T, u <= v\}$$
$$v_0 = Sup\{v \mid (u_0, v) \in T\}$$
$$v_1 = Sup\{v \mid (u, v) \in T, v < u\}$$
$$u_1 = Sup\{u \mid (u, v_1) \in T\}$$

$P^{*'}{}_S(A|B)$ is the only possitive root of the equation
$$ax^2 + bx + c = 0$$
where
$$a = v_1 - v_0$$
$$b = u_0 v_0 - u_1 v_1$$
$$c = (u_1 - u_0) v_0 v_1$$

If the points of $S$ are ordered in lexicographical order (complexity of doing it, $O(n \log(n))$) then all the steps can be carried out in linear time. Then the complexity of the algorithm can be considered $O(n.\log(n))$, where $n$ is the number of points of $S$.

Applying this algorithm to Example 3, we obtain the following intervals:

- For $u_1$ given $r$: $\approx [0.9688, 1]$
- For $u_2$ given $r$: $\approx [0, 0.0312]$

For example 4, we get the same intervals that using $Nextr(\mathcal{P}_B'')$. In its modified version, with probabilities given by expressions (53) and (54), the intervals are:

- For $r$ given $\{r, w\}$: $[0.2929, 0.5]$
- For $w$ given $\{r, w\}$: $[0.5, 0.7071]$

We do not find qualitative differences between the intervals obtained by using this integral and the ones obtained with Choquet's integral. We think that the behaviour (considering the complete set $\mathcal{P}_B$) is appropriate in both cases.

## 5  CONCLUSIONS

In this paper we have studied several methods of conditioning convex sets of probabilities. It has been considered more appropriate the conditioning of all the possible probability distributions and the assignation, at the same time, of a possibility value to each conditional probability. This possibility is calculated taking into account the normalization carried out in its calculation.

We have considered also the problem of assigning an uncertainty interval to each event taking into account the conditional probabilities and its associated possibilities. We have called it uncertainty interval because it is determined taking into account both types of information: the probabilities and the possibilities. It would not be correct to call it possibility (or probability) interval. It is a mixture of possibilities and probabilities. For this task, several methods have been proposed based on the use of Choquet's integral and Sugeno's integral. We do not consider that there are important differences on the use of both integrals. Both have also a similar complexity in its calculus. Sugeno's integral is very easy to interpret: to calculate the maximum value of conditional probabilities, the value of the conditional probability is limited by its possibility. If $P_i(A|B) = 0.9$ and $\pi(p_i(.|B)) = 0.3$, then we only can use 0.3 to calculate the maximum. We have shown that a previous method (Moral, Campos, 1991) based on the consideration of the extreme conditional probabilities reduces too much the length of the intervals.

Finally, we want to point out that the stablished link between possibilities and upper and lower probabilities can be a good basis for an integration of these two theories and deverses future studies.

### Acknowledgements

This work has been supported by the Commission of the European Communities under ESPRIT BRA 3085: DRUMS.